\documentclass[aps,pra,notitlepage,showpacs,superscriptaddress,floatfix,10pt]{revtex4-1}
\usepackage{framed}
\usepackage{graphicx}
\usepackage{amsmath}
\usepackage{epsfig}
\usepackage{helvet}
\usepackage{amssymb}
\usepackage{framed}
\usepackage{bbold}
\usepackage{subfigure}
\usepackage{color}
\usepackage{listings}
\allowdisplaybreaks[3]

\begin{document}
\title{TensorNetwork for Machine Learning}

\author{Stavros Efthymiou}
\affiliation{Alphabet (Google) X, Mountain View, CA 94043, USA}
\affiliation{Max-Planck-Institut f{\"u}r Quantenoptik, Hans-Kopfermann-Strasse 1, 85748 Garching, Germany}

\author{Jack Hidary}
\affiliation{Alphabet (Google) X, Mountain View, CA 94043, USA}

\author{Stefan Leichenauer}
\affiliation{Alphabet (Google) X, Mountain View, CA 94043, USA}

\begin{abstract}
We demonstrate the use of tensor networks for image classification with the TensorNetwork~\cite{library} open source library. We explain in detail the encoding of image data into a matrix product state form, and describe how to contract the network in a way that is parallelizable and well-suited to automatic gradients for optimization. Applying the technique to the MNIST and Fashion-MNIST datasets we find out-of-the-box performance of 98\% and 88\% accuracy, respectively, using the same tensor network architecture. The TensorNetwork library allows us to seamlessly move from CPU to GPU hardware, and we see a factor of more than 10 improvement in computational speed using a GPU.
\end{abstract}

\maketitle


\section{Introduction}

Tensor networks have seen numerous applications in the physical sciences~\cite{Fannes, White, Vidal, Perez-Garcia, MERA, MERA2, MERAalgorithms, Shi, Tagliacozzo, Murg, PEPS1, PEPS2, PEPS3, rev1, rev2, rev3, rev4, rev5, MERA, MERA2, MERAalgorithms, QC1, QC2, QC3, QC4,cMPS,cMERA, CTMRG, TRG, TEFRG, TNR, Swingle, dS1, dS2, dS3, MERAgeometry}, but there has been significant progress recently in applying the same methods to problems in machine learning~\cite{ML1, ML3, ML9, ML2, ML4, ML5, ML6, ML7, ML8, ML10, ML11}. The TensorNetwork library~\cite{library} was created to facilitate this research and accelerate the adoption of tensor network methods by the ML community. In a previous paper~\cite{paper1} we showed how TensorNetwork could be used in a physics setting. Here we are going to illustrate how to use a matrix product state (MPS) tensor network to classify MNIST and Fashion-MNIST images. The basic technique was applied to the MNIST dataset by Stoudenmire and Schwab~\cite{ML1}, who adapted the DMRG algorithm from physics~\cite{White} to train the network. Our implementation differs from theirs and follows more closely the implementation in the specialized TorchMPS library~\cite{torchmps}. The most significant change from Stoudenmire and Schwab is that we use automatic gradients to train the model rather than DMRG. This method of training is familiar to ML practitioners and is built-in to TensorFlow~\cite{TensorFlow}, which is the TensorNetwork backend we use. We also empirically find it useful to use an alternative contraction order compared to Stoudenmire and Schwab when computing components of the MPS for ease of parallelization. MPS methods were previously applied into Fashion-MNIST in~\cite{fashionMPS}, and we achieve the same accuracy here. In terms of speed, we note that a factor of more than 10 improvement is gained moving from CPU to GPU hardware using the same code with a TensorFlow backend.

Up to a few tweaks of implementation, the main strategies we employ for image classification can be found elsewhere. The purpose of this note is to be a resource for machine learning practitioners who wish to learn how tensor networks can be applied to classification problems. An important part of this is the code that we have uploaded to GitHub~\cite{library}, which we hope will also be a valuable resource for the community.

To summarize the results, we find that using an MPS tensor network for classification results in 98\% test accuracy for MNIST and 88\% test accuracy for Fashion-MNIST. The only hyperparameter in the model is the bond dimension, $\chi$, but we find that the results are largely independent of $\chi$ for $\chi \gtrsim 10$. We also compare training times on CPU and GPU, and find that the GPU leads to about 10x speedup over the 64-core CPU.


\section{Setup}

\subsection{Encoding Data in a Tensor Network}\label{sec-encoding}

In this section we will briefly review the structure of a tensor network and how it is used to encode image data. See~\cite{library} for further background that does not assume expertise in quantum physics.

The motivation for using a tensor network for data analysis is similar to the motivation for using a kernel method. Even though tensor networks are most useful for linear operations on data, they act on a very high-dimensional space, and that high dimensionality can be leveraged into representation power. We begin by carefully defining that high-dimensional space.

For us an image consists of a list of $N$ greyscale pixels. In the MNIST and Fashion-MNIST datasets $N = 28^2 = 784$. By flattening the images into a list we lose the two-dimensional nature of the image. This is obviously a drawback and will negatively impact the performance, but our goal is just to illustrate the ideas of a tensor network in a simple application. More complicated schemes could be employed to get around this limitation. See, for example,~\cite{ML9} for an example of two-dimensional tensor network applied to image classification.

The $N$ pixel values of an image will be encoded in a $2^N$-dimensional vector space as follows. First, each pixel of the image is encoded into its own two-dimensional pixel space according to a \textit{local feature map}. One local feature map, used in~\cite{ML1}, is
\[
\Phi(p) = \begin{pmatrix} \cos \pi p/2 \\ \sin{\pi p/2} \end{pmatrix}.
\]
Here $p\in [0,1]$ is the pixel value normalized to unit range. Notice that the pixel values $0$ and $1$ get mapped to independent vectors in this space, and all other pixel values are linear combinations of those. If the image were purely black-and-white this would simply be a one-hot encoding of the pixel values. Another feature map which has similar properties is
\[
\Phi(p) = \begin{pmatrix} 1-p \\ p \end{pmatrix}.
\]
In practice we will use the latter, linear feature map, though this is not very important. With color images it would make sense to consider a higher-dimensional pixel space, such as a $2\times 3 = 6$-dimensional space encoding the RGB values of each pixel. Finally, by way of notation we will refer to the components $\Phi(p)_i$, where the $i$ index takes on two values. So for the linear feature map, $\Phi(p)_0 = 1-p$ and $\Phi(p)_1 = p$.

The total image space is defined as the tensor product of all of the pixel spaces. A key property of this space is that flipping a single pixel value from black to white results in an independent image vector. In equations, an image $(p_1,p_2,\ldots, p_N)$ is encoded in the image space as
\[
(p_1,p_2,\ldots, p_N) \mapsto \Phi(p_1) \otimes \Phi(p_2) \otimes \cdots \otimes \Phi(p_N).
\]
We will refer to this object as a ``data tensor" or ``data state." The data tensor has $2^N$ components, each of which is the product of one of the two components of the local feature map (e.g., $p$ or $1-p$) over all of the pixels. In index notation, the components of the data tensor are given by
\[
\Phi(p_1)_{i_1} \Phi(p_2)_{i_2} \cdots \Phi(p_N)_{i_N}.
\]
Since there are two possible values for each of the $i_k$ indices, the total number of components is $2^N$.

\begin{figure}
    \includegraphics[width=0.4\textwidth]{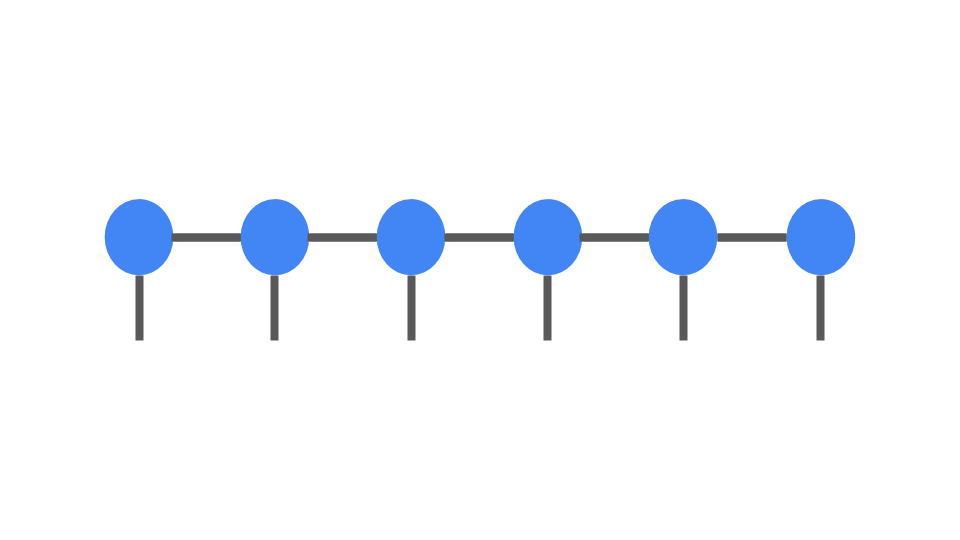}
    \caption{Matrix Product State (MPS) tensor network. The free indices correspond to the pixels in the image.}\label{fig-MPS}
\end{figure}

The MPS tensor network represents another type of vector in the image space. As such, it also has $2^N$ components, and the MPS writes each of those components in terms of the product of $N$ matrices. Letting $T$ represent the total MPS tensor, we have
\begin{equation}\label{eq-MPS}
T_{i_1i_2\cdots i_N} = \sum_{\alpha_1,\alpha_2,\ldots\alpha_N} A^{(1)}_{i_1 \alpha_1}A^{(2)}_{i_2 \alpha_1\alpha_2}A^{(3)}_{i_3 \alpha_2\alpha_3}\cdots A^{(N)}_{i_N \alpha_N}.
\end{equation}
The ranges of each of the $\alpha_k$ indices, called the bond dimensions, are hyperparameters of the model. The bond dimensions determine the sizes of the $A$ tensors. The components of the $A$ tensors are variational parameters, or weights, that are fixed via training. There is some redundancy in those parameters, known as gauge freedom~\cite{gauge}, but we will not concern ourselves with that here.  

To summarize, we have described an image space and an embedding of our data into that space in the form of the data tensors. The MPS tensor $T$ is another vector in that space which is not itself the data tensor of a single image. Intuitively speaking, we would like the MPS tensor $T$ to be equal to a linear combination of all of the images in a given class. An image not belonging to the class will be orthogonal to $T$, while an image belonging to the class will not. In the next section we will describe how to train the MPS to have this property. When we have multiple classes to label, we can either construct MPS vectors or add an extra ``label" node to the MPS to keep track of the class. The details of this are discussed in the following sections.


\subsection{Objective Function}

The classification task can be expressed as finding a map $f$ from the space of objects to the space of labels. In the MNIST case $f$ should map each handwritten image to the corresponding digit from the set $\{0,1,2,\dots ,9\}$. In a machine learning setting $f$ is parametrized using a large number of variational parameters that are then optimized using pairs $(\mathbf{x},y)$ of labeled examples. Here $\mathbf{x}=(p_1,p_2,\dots p_N)\in [0,1]^N$ represents a flattened image and $y\in \{0, 1, \dots, L-1\}$ the corresponding label (with $N=784$ and $L=10$ for MNIST). Typical choices for such parametrization range from a simple linear regression or support vector machines to more complicated deep neural networks. In our setting, following~\cite{ML1}, we define the classification map as follows: First we calculate the inner product between the encoded image vector (see Section~\ref{sec-encoding}) and a variational MPS $T_{i_1i_2...i_N}^l$:
\begin{figure}
    \includegraphics[width=0.4\textwidth]{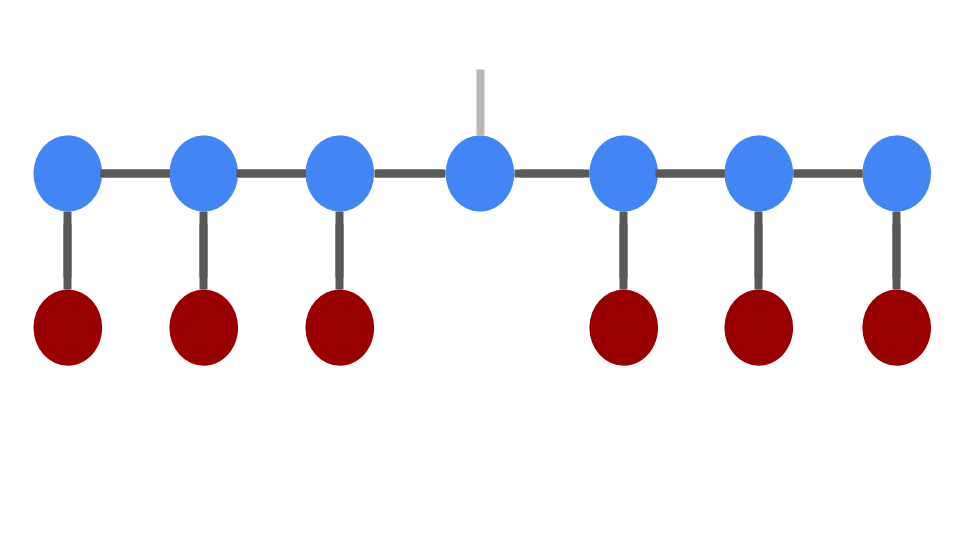}
    \caption{Inner product between the variational MPS (blue nodes) and the encoded data vector (red nodes). Notice the free MPS label index, depicted in lighter grey.}\label{fig-MPS-data}
\end{figure}

\begin{equation}\label{eq-f}
f^{(l)}(\mathbf{x}) = \sum _{i_1,i_2,...,i_N=0}^1T_{i_1i_2...i_N}^l\Phi(p_1)_{i_1} \Phi(p_2)_{i_2} \cdots \Phi(p_N)_{i_n}.
\end{equation}
The inner product is depicted in Fig.~\ref{fig-MPS-data} in tensor network graphical notation. Note that all pixel indices of the MPS are contracted with data, except the index $l=0,1,\dots, L-1$ which is free and used to distinguish the different labels. The position of the $l$ index in the MPS chain is arbitrary, and a typical choice is in the middle (position $N/2$). After calculating the inner product $f^{(l)}(\mathbf{x})$ the classification map is defined as:
\[
f(\mathbf{x}) = \mathrm{argmax}_l~f^{(l)}(\mathbf{x}).
\]
In other words, for each image we select the label whose MPS has the largest overlap with the corresponding encoded image vector.

Following the typical machine learning procedure, the variational parameters $A^{(j)}_{i_j\alpha \beta }$ that define each MPS $T$ (see Eq.~\eqref{eq-MPS}) should be tuned to minimize an objective function in the training set. In the original work~\cite{ML1}, the average mean squared error was chosen as the objective function. Here we choose instead to optimize the multi-class cross-entropy defined on our training set $\mathcal{D}$ as:
\begin{equation}\label{eq-loss}
\text{CE} = -\sum _{(\mathbf{x}_i, y_i)\in \mathcal{D}} \log \mathrm{softmax}~f^{(y_i)}(\mathbf{x}_i),
\end{equation}
where
\[
\mathrm{softmax}~f^{(y_i)}(\mathbf{x}_i) = \frac{e^{f^{(y_i)}(\mathbf{x}_i)}}{\sum _{l=0}^{L-1}e^{f^{(l)}(\mathbf{x}_i)}}.
\]
Note that the outputs the softmax function can be interpreted as the predicted probabilities for each label. The final prediction corresponds to the label with the maximum probability. The cross-entropy following a softmax activation is a choice that is well suited to classification problems, as is known from standard machine learning methods.


\subsection{Implementation}
High level frameworks such as TensorFlow allow for an efficient implementation of standard machine learning algorithms, such us neural networks, through a very simple API. One of the reasons for this simplicity is the automatic calculation of gradients used in the famous backpropagation algorithm~\cite{backprop}. The user only needs to define the forward pass of the model, which is generally straightforward, while the more complicated backward pass is handled automatically by the library. 

Gradient optimization methods are not the typical choice for optimizing tensor networks. In most physics applications, sweeping algorithms, such as the celebrated Density Matrix Renormalization Group (DMRG)~\cite{White} are preferred, as they can lead to faster convergence. Regardless of that, a "brute force" optimization using the gradients of the objective function is will work for the tensor network case, too. This approach may be suboptimal when compared to a more sophisticated sweeping method, but the simplicity of the underlying code for gradient-based optimization when written in a high-level machine learning library can be more attractive to machine learning practitioners.
\begin{figure}
    \includegraphics[width=0.7\textwidth]{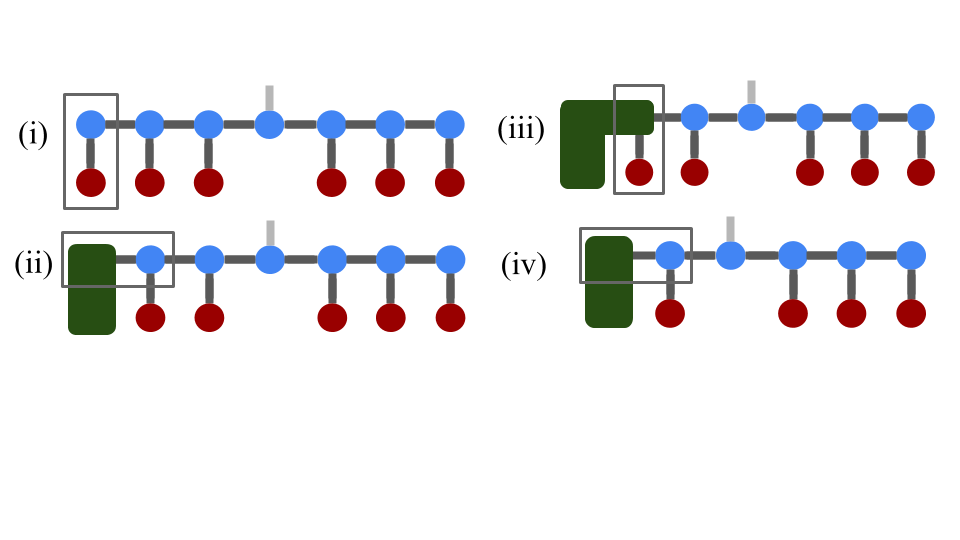}
    \caption{Contract MPS in order.}\label{fig-MPS-left-contr}
\end{figure}

The way the automatic differentiation works is strongly dependent on how the forward pass is defined by the user. When using high-level libraries, one should take advantage of the efficiency of vectorized operations, as this will lead to a more efficient forward pass and possibly more efficient gradients as well. In our case, the forward pass amounts to calculating the inner product of Eq.~\eqref{eq-f} (see Fig.~\ref{fig-MPS-data}).

A straightforward way to calculate this inner product (or equivalently to contract the tensor network), which is also commonly used in sweeping methods, is depicted in Fig.~\ref{fig-MPS-left-contr}. Denoting with $d$ the dimension of the pixel space (size of vertical legs in the figure, where $d=2$ for the feature map described in Section~\ref{sec-encoding}) and $\chi $ the bond dimension (size of horizontal legs) which we assume to be constant across the MPS, then the cost of contracting a vertical leg is $O(\chi d)$, while the cost of contracting a horizontal leg is $O(d\chi ^2)$. Assuming that we continue the contraction as depicted from left to right, once we pass the free label index we will have to keep tack of the label for the rest of the contractions, increasing all costs by a factor of $L$ (the number of different labels) and leading to a total cost of order $O(NLd\chi ^2)$. An easy way to avoid the extra factor of $L$ is to start a contraction from both ends of the chain and contract with the tensor that carries the label index in the final step, resulting to an improved total cost of $O(Nd\chi ^2)$. Note that this analysis only takes into account the forward pass, that is calculating $f^{(l)}(\mathbf{x})$ for a given $\mathbf{x}$, and not its gradients with respect the MPS parameters, for which we cannot avoid the additional $L$ factors.
\begin{figure}
    \includegraphics[width=0.7\textwidth]{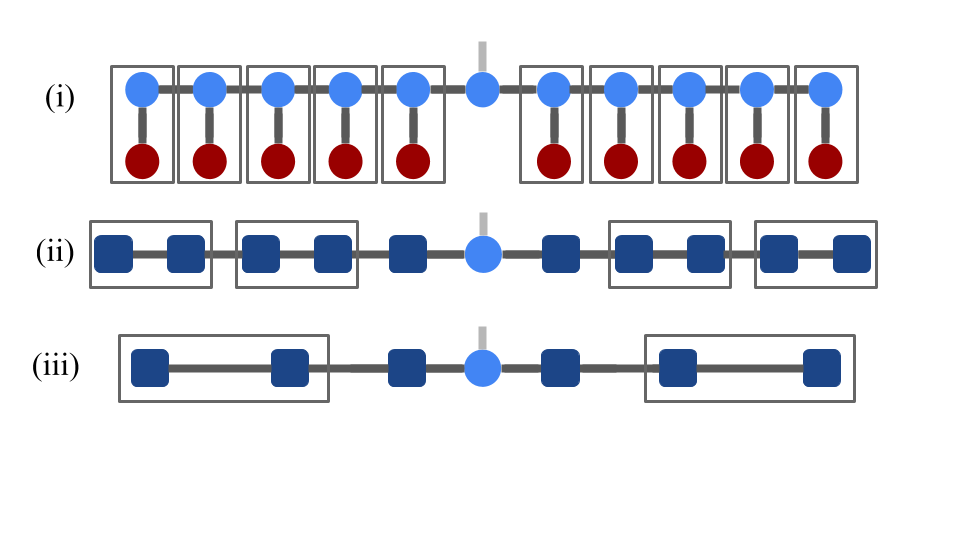}
    \caption{Parallelized MPS contraction. (i) Contract all of the pixel indices with a data tensor. This creates new effective tensors depicted as dark blue squares. (ii) Contract the new tensors in pairs. This step can be done independently and in parallel for each pair. The result is a new chain with half as many effective tensors. (iii) Contract again in pairs, and repeat until the chain is fully contracted.}\label{fig-MPS-bin-contr}
\end{figure}

Although this method of contraction is expected to work well when coded using a low-level language, we empirically find it to be suboptimal for our specific application and the particular choices of the parameters $N$, $d$ and $\chi $, both in the forward pass and also in the automatic backward pass. We follow an alternative contraction order inspired by the implementation of~\cite{torchmps} and depicted in Fig~\ref{fig-MPS-bin-contr}. The total cost of this contraction is $O(Nd\chi ^2 + \chi^3\log N)$, where the first term comes from step (i) and the second term from the consecutive pairwise contractions. This method requires fewer contractions per site, however the cost scales as $\chi^3$ as it requires matrix-matrix multiplications. In contrast, the first method only has matrix-vector contractions. Even though the total cost is higher asymptotically for the second method, an advantage is that each step is easy to parallelize as the matrix multiplications are independent and do not require results from the neighboring calculation (as they do in the first method). This is of particular importance when using a machine learning library such as TensorFlow, as the supported batching operations can be used to easily implement these contractions in parallel. We note that this implementation does not only lead to a faster forward pass, but also to a more efficient automatic gradient calculation.


\subsection{Optimization}
As described in the previous section, upon defining our model's forward pass as depicted in Fig.~\ref{fig-MPS-bin-contr}, TensorFlow automatically calculates the gradients of the loss function in Eq.~\eqref{eq-loss}. These gradients are then used to minimize the loss via a stochastic gradient descent method. A typical setting that we find to perform well is to use Adam optimizer~\cite{adam} with a learning rate of $10^{-4}$ and batch sizes ranging from $10$ to $100$ samples depending on the total amount of training data used. We note that in the original sweeping DMRG-like optimization method proposed in~\cite{ML1} each step updates two of the MPS tensors. In contrast, in gradient based methods using automatic differentiation typically all variational tensors are updated simultaneously in each update step.

A disadvantage of this method, when implemented naively, is that the bond dimension $\chi $ is an additional hyperparameter that is set a priori and is kept constant during training. In the sweeping implementation, a singular value decomposition (SVD) step allows to adaptively change bond dimensions during training. This is a particularly interesting feature as it allows the model to change the number of its variational parameters according to the complexity of data to be learned.


\section{Results}
\begin{figure}
    \includegraphics[width=0.9\textwidth]{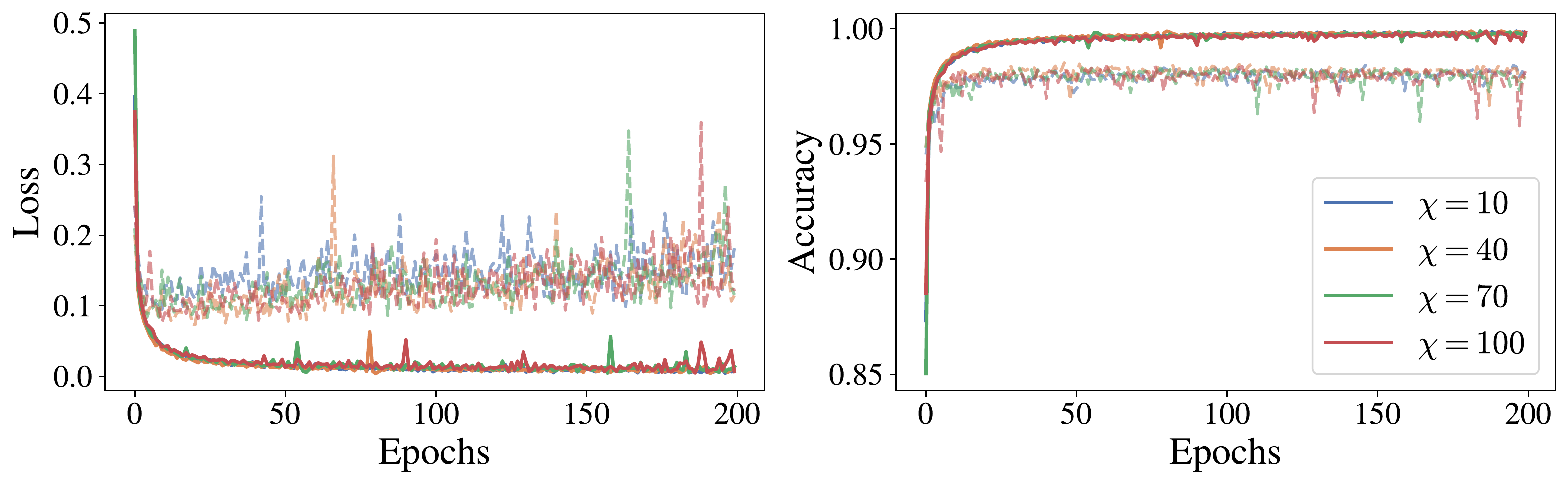}
    \caption{Evolution of training (solid) and test (dashed) loss and accuracy during training on the MNIST dataset.}\label{fig-mnist-training}
\end{figure}
\begin{figure}
    \includegraphics[width=0.9\textwidth]{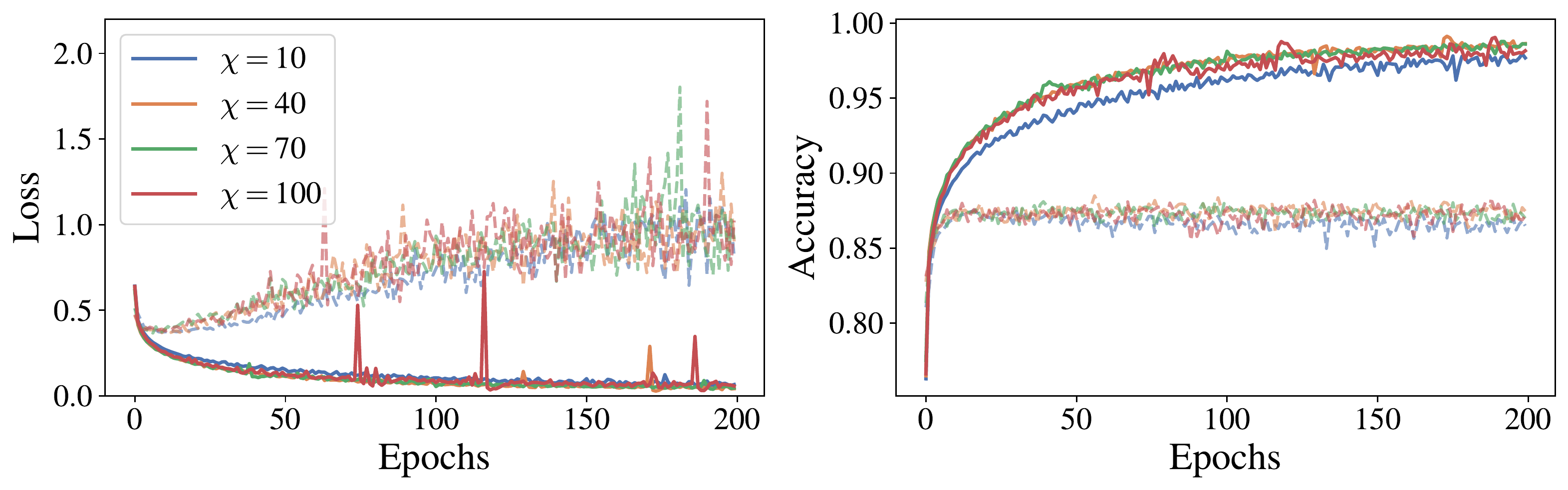}
    \caption{Evolution of training (solid) and test (dashed) loss and accuracy during training on the Fashion-MNIST dataset.}\label{fig-fashion-training}
\end{figure}

We implement the contraction method described in Fig.~\ref{fig-MPS-bin-contr} using TensorNetwork with the TensorFlow backend, and we optimize using the built-in automatic differentiation and Adam optimizer with learning rate set to $10^{-4}$.

\begin{figure}
    \includegraphics[width=0.9\textwidth]{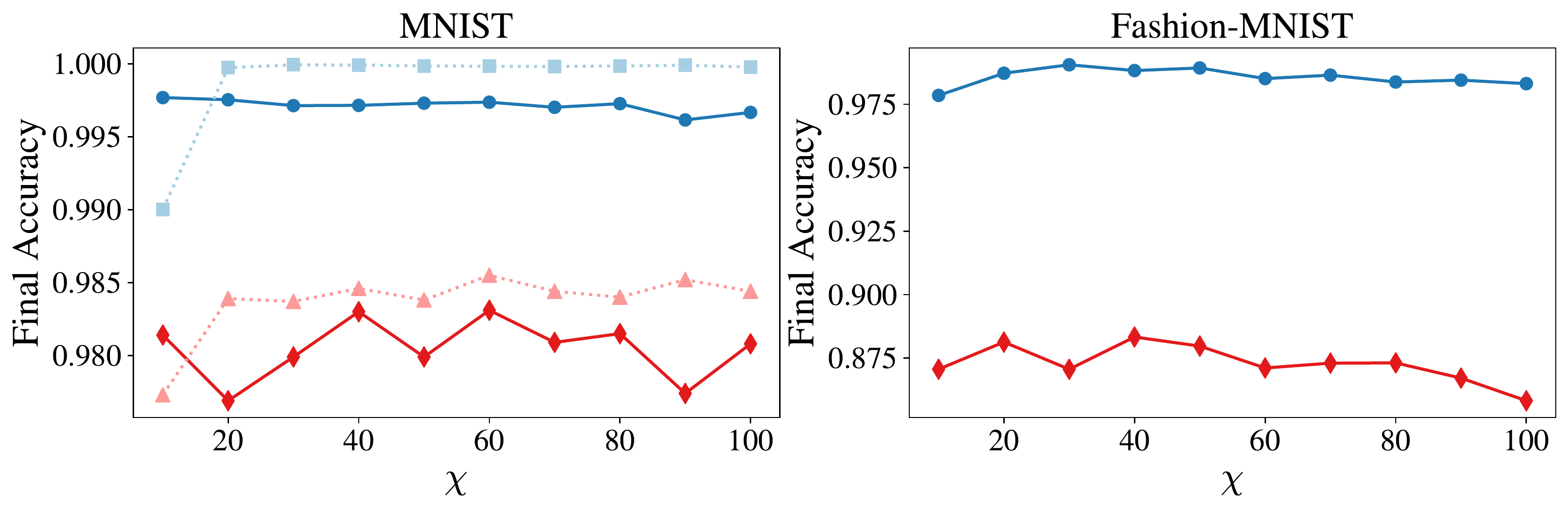}
    \caption{Final accuracies on the full train (60,000 images) and test (10,000) sets as a function of bond dimension. The left plot is on the MNIST dataset of hand-written digits and the right plot on Fashion-MNIST. Blue color (squares and circles) corresponds to the training set and red color (triangles and diamonds) to the test set. For MNIST we also compare performance using the cross-entropy loss (solid lines) to the mean square loss (dotted lines) employed in~\cite{ML1}.}\label{fig-acc}
\end{figure}

We first train on the total MNIST training set consisting of 60,000 images of size $28\times 28$ and we show the relevant training history in Fig.~\ref{fig-mnist-training}. Each training epoch corresponds to a full iteration over the training set using a batch size of $50$. We observe that with this setting the model requires about $50$ epochs to converge to almost 100\% training accuracy and about 98\% test accuracy. Here test accuracy is calculated on the whole test dataset consisting of 10,000 images. Training with automatic gradients is found to be independent of the used bond dimension, with largest bond dimensions being of course more computationally expensive. In Fig.~\ref{fig-acc} left we plot the final training and test accuracies as a function of the bond dimension. Here the test accuracy is calculated on the MNIST test set of $10,000$ images once training is completed. We again find no dependence on the bond dimension obtaining about 98\% test accuracy, in agreement with previous works. Furthermore we compare optimizing with our softmax cross-entropy loss (denoted as CE) with the original mean square loss from~\cite{ML1} and we find no significant difference in terms of final accuracies. The square loss is more efficient to calculate and thus leads to slightly faster convergence in terms of actual wall time, however the difference is negligible as the bond dimension is increased, when contractions dominate the computation time.

We repeat the same optimization schedule for Fashion-MNIST, a dataset that has exactly the same structure as MNIST ($28\times 28$ grayscale images of clothing with $L=10$ labels), however it is significantly harder to classify, with state of the art deep learning methods obtaining about 93\% test accuracy~\cite{fashionSOTA}. The corresponding training dynamics are shown in Fig.~\ref{fig-fashion-training}. As demonstrated in Fig.~\ref{fig-acc} right we are able to obtain 88\% test accuracy and we again find no significant dependence on the bond dimension.

\begin{figure}
    \includegraphics[width=0.7\textwidth]{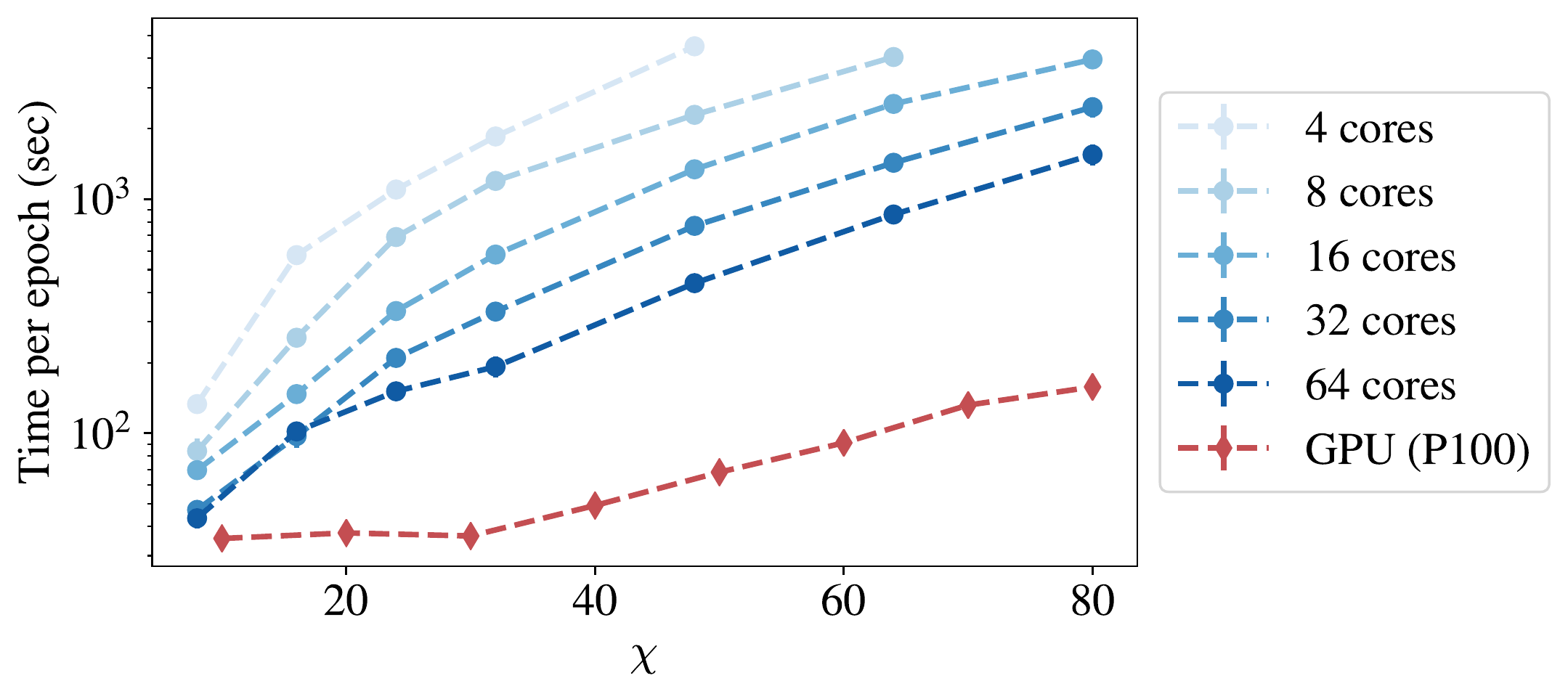}
    \caption{Wall-time required per optimization epoch when implementing the same TensorNetwork/TensorFlow code on GPU and CPU. One epoch corresponds to a full iteration over the whole training set.}\label{fig-times}
\end{figure}

Finally, it is well known that the use of accelerators such us GPU can greatly reduce training time for many machine learning models, particularly in cases where complexity is dominated by linear algebra operations. Since this is the case with tensor networks, we expect to see some advantage in our methods. TensorNetwork with the TensorFlow backend allows direct implementation on a GPU without any changes in the code. In Fig.~\ref{fig-times} we verify the advantage, with the GPU being four times faster for the smallest bond dimension and the relative speed-up increasing with increasing bond dimension.


\section{Conclusion}

Here we have described a tensor network algorithm for image classification using MPS tensor networks. All of the code required for reproducing the results in the open source TensorNetwork library is available on GitHub~\cite{library}. We are hopeful that the techniques we reviewed here will be taken up by the wider ML community, and that the code examples we are providing will become a valuable resource. By using TensorFlow as a backend, we were able to access automatic gradients for optimization of the tensor network, which moves us beyond the physics-centric techniques ordinarily used. Furthermore, TensorFlow already offers high-level methods for deploying state-of-the-art deep learning models. Adding tensor network machinery to that same library will allow direct comparison between these tools to more traditional machine learning methods, such as neural networks, a particularly active research area. Ultimately, one might be even able to further push the state-of-the-art by combining tensor networks with more traditional methods, something that can be implemented very easily using TensorNetwork on top of TensorFlow.

\begin{acknowledgements}
We would like to thank Glen Evenbly, Martin Ganahl, Ash Milsted, Chase Roberts, Miles Stoudenmire, and Guifre Vidal for valuable discussions. X is formerly known as Google[x] and is part of the Alphabet family of companies, which includes Google, Verily, Waymo, and others (www.x.company).
\end{acknowledgements}

\end{document}